\documentclass[sigconf]{acmart}

\AtBeginDocument{%
  \providecommand\BibTeX{{%
    \normalfont B\kern-0.5em{\scshape i\kern-0.25em b}\kern-0.8em\TeX}}}

\copyrightyear{2022}
\acmYear{2022}
\setcopyright{acmlicensed}\acmConference[ICONS 2022]{International Conference on Neuromorphic Systems}{July 27--29, 2022}{Knoxville, TN, USA}
\acmBooktitle{International Conference on Neuromorphic Systems (ICONS 2022), July 27--29, 2022, Knoxville, TN, USA}
\acmPrice{15.00}
\acmDOI{10.1145/3546790.3546824}
\acmISBN{978-1-4503-9789-6/22/07}

\begin{document}

\title{Neuro-symbolic computing with spiking neural networks}

\author{Dominik Dold}
\email{dominik.dold@esa.int}
\affiliation{%
  \institution{European Space Agency (ESTEC),\\Advanced Concepts Team}
  \city{Noordwijk}
  \country{Netherlands}
}

\author{Josep Soler Garrido}
\email{josep.soler-garrido@ec.europa.eu}
\affiliation{%
  \institution{European Commission,\\Joint Research Centre (JRC)}
  \city{Seville}
  \country{Spain}
}

\author{Victor Caceres Chian}
\email{vcacereschian@gmail.com}
\affiliation{%
  \institution{Siemens AG Technology}
  \city{Munich}
  \country{Germany}
}

\author{Marcel Hildebrandt}
\email{marcel.hildebrandt@siemens.com}
\affiliation{%
  \institution{Siemens AG Technology}
  \city{Munich}
  \country{Germany}
}

\author{Thomas Runkler}
\email{thomas.runkler@siemens.com}
\affiliation{%
  \institution{Siemens AG Technology}
  \city{Munich}
  \country{Germany}
}

\begin{abstract}
Knowledge graphs are an expressive and widely used data structure due to their ability to integrate data from different domains in a sensible and machine-readable way.
Thus, they can be used to model a variety of systems such as molecules and social networks.
However, it still remains an open question how symbolic reasoning could be realized in spiking systems and, therefore, how spiking neural networks could be applied to such graph data.
Here, we extend previous work on spike-based graph algorithms by demonstrating how symbolic and multi-relational information can be encoded using spiking neurons, allowing reasoning over symbolic structures like knowledge graphs with spiking neural networks.
The introduced framework is enabled by combining the graph embedding paradigm and the recent progress in training spiking neural networks using error backpropagation.
The presented methods are applicable to a variety of spiking neuron models and can be trained end-to-end in combination with other differentiable network architectures, which we demonstrate by implementing a spiking relational graph neural network.
\end{abstract}



\maketitle

\section{Introduction}

How information is encoded in the temporal domain of spikes is still an open question \cite{zenke2021visualizing}.
Even though there has been a lot of progress in applying spiking neural networks (SNNs) to sensory machine learning tasks (e.g., visual and auditory information processing) \cite{zenke2018superspike,mostafa2017supervised,comsa2019temporal,kheradpisheh2019s4nn,goltz2021fast,wunderlich2021event,yin2021accurate}, there is a lack of work on using SNNs on relational data and reasoning tasks \cite{crawford2016biologically,dold2021spikeembed,chian2021learning,dold2022spikeembed}.
Here, we present a method that enables learning of spike-based representations of abstract concepts and their relationships, which can subsequently be used to reason in the underlying semantic space.

\begin{figure}[ht!]
  \centering
    \includegraphics[width=\columnwidth]{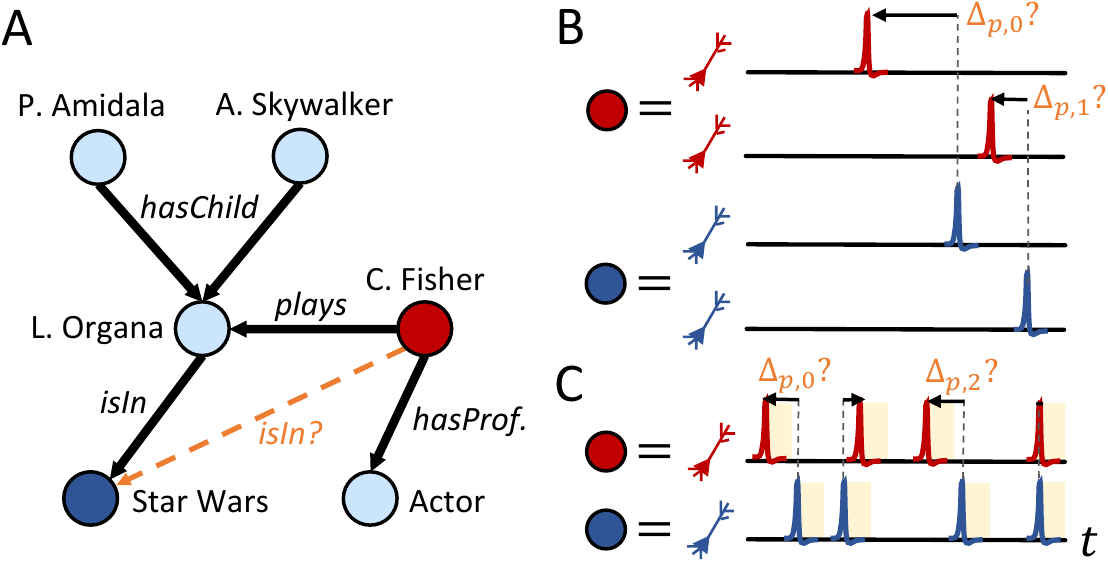}
    \Description{}
    \caption{\textbf{(A)} Example of a KG containing information about actors and movies.
    Entities are represented by nodes (circles), while relations of different types are represented by edges (arrows).
    Using the information available in the KG, the existence of novel edges can be predicted (link prediction, dashed line).
    \textbf{(B)} Time to first spike embeddings of graph elements, where each entity in the graph is encoded by the times to first spike of a population of neurons.
    Relations $p$ are embedded as spike time differences.
    A relation, e.g. '\textit{isIn}' from (A), is considered to hold between two entities if the spike time differences between the entity-embedding populations match the relation embedding.
    \textbf{(C)} Spike train embeddings of graph elements, where each entity is represented by a single neuron.
    In this case, neurons become refractory (yellow shade) after spiking.
    Figures adapted from \cite{dold2022spikeembed}.}\label{fig}
\end{figure}

Our approach is based on knowledge graphs (KGs) \cite{auer2007dbpedia,bollacker2008freebase,singhal2012introducing} and knowledge graph embedding algorithms \cite{nickel2015review,hamilton2017representation,ruffinelli2019you}, extending previous work on spike-based algorithms for homogeneous graphs \cite{hamilton2017community,hamilton2018towards,hamilton2018neural,schuman2019shortest,ali2019spiking,hamilton2019spike,kay2020neuromorphic}.
KGs enable a structured organization of information from different domains and modalities.
One particular simple description of KGs is as a list of triple statements (subject $s$, predicate $p$, object $o$) stating that a relationship $p$ holds between two entities $s$ and $o$ \cite{brickley1999resource}.
An equivalent representation is as a graph, where a triple corresponds to a link in the graph, i.e., an edge of type $p$ connecting two nodes representing entity $s$ and $o$ (Fig.~\ref{fig}A).
Using the information contained in a KG, one can reason about the validity of novel triple statements (link prediction task) or about the relatedness of entities (node classification task).
One way of performing such reasoning tasks is by extracting rules (i.e., a certain pattern of links in the graph that is true for several entities) from the KG (see, e.g., \cite{lehmann2009dl}) -- similar to constructing filters like edge detectors for image recognition tasks.
However, especially for large KGs with many facts, which are usually incomplete and can contain wrong statements as well, finding appropriate rules becomes hard.

A widely adopted method for inference on KGs is knowledge graph embedding, where elements of the graph are mapped into a low-dimensional vector space while conserving certain graph properties.
The learned vector representations contain contextual information from the KG and can consequently be used either directly or in combination with machine learning methods like artificial neural networks to perform inference tasks.
We explore a similar approach to learn spike-based embeddings for KGs, (i) making data organized as KGs accessible to the temporal world of SNNs and (ii) demonstrating how neuro-symbolic computations can be realized in spike-based systems.
The discussed results have been recently published in \cite{dold2021spikeembed,chian2021learning,dold2022spikeembed}.

\section{Spike-based graph embeddings}

We found that one way of embedding graph elements in the temporal domain of spikes is by representing entities as the time to first spike of neuron populations and relations as spike time differences between populations \cite{dold2021spikeembed}.
We further extended this coding scheme in two ways: (i) instead of using neuron populations, each entity can be represented by the spike train of a single neuron, fully utilizing the temporal domain \cite{dold2022spikeembed} and (ii) we demonstrated in a proof-of-concept that this coding scheme can be used to build spiking graph neural networks (GNNs), introducing an event-based, deep and sparse embedding algorithm suitable for various KG inference tasks \cite{chian2021learning}.
In the latter case, we further found that the convolutional (and hence, non-locally shared) weights encountered in modern GNNs\footnote{Modern GNN architectures extend the convolution operator used in convolutional neural networks (CNNs) for images to graphs. See Fig.~\ref{figGNN}A,B for a detailed description.} can be frozen, increasing the compatibility of the proposed algorithm with neuromorphic architectures that often adapt a distributed and locally-constrained design philosophy \cite{frenkel2021bottom}.

For all these models, spike-based embeddings are found using gradient-based optimization on a cost function that encodes some property of interest in the KG, e.g., the likelihood that a link exists in the KG.
Thus, even though we only describe our approach for one specific neuron model in the following, the presented methods can be applied to arbitrary spiking models, as long as gradients can be calculated with respect to spike times -- something that is often guaranteed nowadays due to the success of the surrogate gradient method \cite{zenke2018superspike,neftci2019surrogate}.

\subsection{Time to first spike embeddings}\label{sec:ttfs}

We propose to represent a node $n$ in a KG by the first spike times $\pmb t_{n} \in \mathbb{R}^N$ of a population of $N \in \mathbb{N}$ integrate-and-fire neurons with an exponential synaptic kernel $\kappa(x,y) = \theta\left(x-y\right) \exp\left(-\frac{x-y}{\tau_\mathrm{s}}\right)$,
\begin{equation}\label{eq:dotu}
    \frac{\mathrm{d}}{\mathrm{d}t}u_{n,i}(t) = \frac{1}{\tau_\mathrm{s}}\sum_{j} w_{n,ij} \, \kappa(t, t^\mathrm{I}_j) \,,
\end{equation}
where $u_{n,i}$ is the membrane potential of the $i$th neuron of population $n$, $\tau_\mathrm{s}$ the synaptic time constant and $\theta\left(\cdot\right)$ the Heaviside function.
A spike is emitted when the membrane potential crosses a threshold value $u_\mathrm{th}$.
Since neurons spike only once here, no refractory periods are modelled.
$w_{n,ij}$ are synaptic weights from a pre-synaptic neuron population, with every neuron $j$ emitting a single spike at fixed time $t^\mathrm{I}_j$.
Relations are encoded by a $N$-dimensional vector of spike time differences $\pmb{\Delta}_p \in \mathbb{R}^N$.
The validity of a triple $(s,p,o)$, i.e., (subject $s$, predicate $p$, object $o$) or (entity $s$, relation type $p$, entity $o$), is evaluated based on the discrepancy between the spike time differences of the node embeddings, $\pmb{t}_{s} - \pmb{t}_{o}$, and the relation embedding $\pmb{\Delta}_p$ (Fig.~\ref{fig}B)
\begin{equation}\label{eq:spikedecoder}
    d(s,p,o) = \| \pmb{t}_{s} - \pmb{t}_{o} -\pmb{\Delta}_p \| \,,
\end{equation}
where $\left\lVert\cdot\right\rVert$ is the L1 norm.
If a triple is valid, then the patterns of node and relation embeddings match, leading to $d(s,p,o) \approx 0$, i.e., $\pmb{t}_{s} \approx \pmb{t}_{o} + \pmb{\Delta}_p$.
If the triple is not valid, we have $d(s,p,o) > 0$, with higher discrepancies representing a lower validity score.
Given a KG, such spike embeddings are found using gradient descent-based optimization with the objective of reconstructing the initial KG.

The proposed embedding method reaches similar performance levels than standard graph embedding algorithms on benchmark data sets.
For instance, on the biomedical knowledge graph UMLS \cite{mccray2003upper}, a mean reciprocal rank (MRR) of $0.78$ is reached, compared to $0.81$ reached by the algorithm TransE \cite{bordes2013translating}.
MRR is a metric used to evaluate the performance of an algorithm on the link prediction task, which takes values between $0$ and $1$ (with $1$ being the best, see \cite{ruffinelli2019you} for a more thorough introduction).

\subsection{Spike train embeddings}

Alternatively, instead of using neuron populations, each node can be represented by the spike train of a single neuron. 
In this case, each neuron is allowed to spike more than once and hence, $t_{n,i}$ represents the $i$'th spike of the neuron\footnote{I.e., the spike train embedding $\pmb{t}_{n}$ is given by a sorted vector of spike times.} representing entity $n$,
\begin{equation}
    t_{n,i} = \sum_{j\leq i} I_{n,j} + i \cdot \tau_\mathrm{ref} \,,
\end{equation}
where $I_{n,j}$ are the neuron's interspike intervals (obtained using Eq.~\ref{eq:dotu}) and $\tau_\mathrm{ref}$ its refractory period\footnote{More specifically, $I_{n,j}$ is the time to spike of neuron $n$ after recovering from the refractory period. Thus, $I_{n,j}$ does not include the refractory period itself yet.}.
The embedding for relation $p$, $\Delta_{p,i}$, now encodes the expected spike time difference between the $i$th spike of two node-representing neurons (Fig.~\ref{fig}C).
Using this notation, the validity of triples is again evaluated using Eq.~\ref{eq:spikedecoder}, and embeddings are found by optimizing both the interspike intervals (by modifying the input weights $w_{n,ij}$) and relation embeddings via gradient descent.
This method also reaches competitive performance levels on a variety of benchmarks, e.g., a MRR of $0.78$ on UMLS and a MRR of $0.21$ on the much larger KG FB15k-237 (based on Freebase \cite{bollacker2008freebase}), which is again comparable to the values obtained via TransE.\footnote{However, for FB15k-237 this result was obtained using an abstract spiking neuron model that is faster to simulate, not an integrate-and-fire model, see \cite{dold2022spikeembed} for details.}

\subsection{Spiking graph neural networks}

\begin{figure}[t!]
  \centering
    \includegraphics[width=\columnwidth]{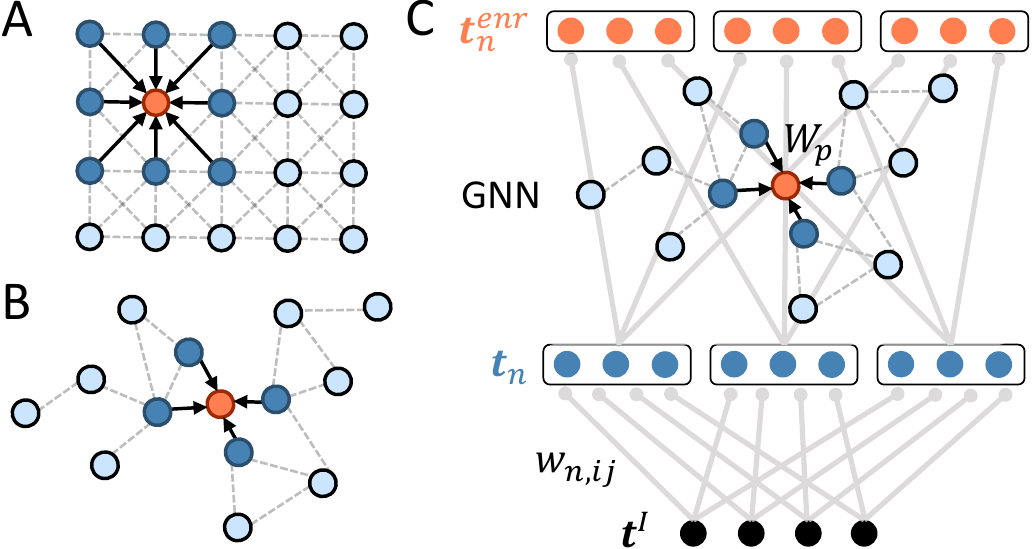}
    \Description{}
    \caption{\textbf{(A)} Images can be represented as graphs with a regular grid structure (dashed lines). In convolutional neural networks, local information (blue) is combined through learnable filters (arrows) to extract features at an image location (orange), which is shifted over the whole image to extract features at different image locations.
    \textbf{(B)} The same logic can be applied to irregular graphs, but the convolution operation has to account for the fact that each node has a different amount of neighbouring nodes whose location (or order) carries no information. 
    \textbf{(C)} Network structure for spike-based graph embeddings with one population (boxes) per node. The network is driven by periodic static random spikes $\pmb{t}^I$. Through the weights $w_{n,ij}$, initial spike embeddings $\pmb{t}_{n}$ are formed, which are subsequently enriched using a GNN layer with relation-specific weights $W_p$ to create the final spike embeddings $\pmb{t}^{enr}_n$. Figure adapted from \cite{chian2021learning}.}\label{figGNN}
\end{figure}

The previously described models only allow us to reason about concepts (i.e., entities) that have been present in the KG during training time.
This can be relaxed by combining the time to first spike coding scheme (Section~\ref{sec:ttfs}) with GNNs \cite{kipf2016semi,hamilton2017inductive,schlichtkrull2018modeling}.
In modern GNNs, node embeddings are further enriched via neighbouring nodes using a convolution operator.
For instance, in Fig.~\ref{figGNN}A,B, the orange node's embedding is updated by aggregating (arrows) the embeddings of its neighbouring nodes (blue).
This is similar to how the same 2D filter is applied pixel by pixel on an image in convolutional neural networks (Fig.~\ref{figGNN}A) -- however, due to the irregular structure of graphs, the convolution operator has to, for example, be invariant under node permutations (Fig.~\ref{figGNN}B).

As long as we know some facts (i.e., triples) about a previously unseen entity, we can calculate its embedding using the GNN and reason about its relationship with other KG entities.
The dynamics of the neurons encoding the enriched embeddings is given by
\begin{equation}
\label{eq:sgnn}
 \frac{\mathrm{d}}{\mathrm{d}t} \pmb{u}_n^\mathrm{enr}(t) = \frac{1}{\tau_\mathrm{s}} \sum_{p \in \mathcal{R}}  \sum_{j \in \mathcal{N}_n^p} \frac{1}{|\mathcal{N}_n^p|} W_p \pmb{\kappa}(t,\pmb{t}_{j}) \,,
\end{equation}
where $\pmb{\kappa}$ is applied component-wise (i.e., $\pmb{\kappa}(x,\pmb{y})_i = \kappa(x, y_i)$) and neurons spike when crossing the threshold $u_\mathrm{th}$.
$\mathcal{N}_n^p$ is a set containing all nodes connected to node $n$ via relation $p$, $|\cdot|$ denotes the number of elements in a set and $\mathcal{R}$ is the set of all relation types.
$W_p$ are learnable, relation-specific weights (for relation type $p$).
These dynamics directly mimic the graph convolution architecture \cite{kipf2016semi,hamilton2017inductive,schlichtkrull2018modeling}: to update the embedding of node $n$, the embeddings of its neighbouring nodes ($j \in \mathcal{N}_n^p$) are linearly mapped via relation-specific weights $W_p$ and averaged.
This operation is applied to all nodes using shared weights ('convolution') to update every node embedding.
The whole architecture is illustrated in Fig.~\ref{figGNN}C.

We found that the weights $W_p$ of the convolution operator can be kept constant (frozen), with the intuition being that the initial spike embeddings align themselves with the weights $W_p$ during training.
This greatly reduces the complexity of our model and avoids synchronously updating the weights in the convolutional layers, which would conflict with the distributed design philosophy of most neuromorphic architectures.

\begin{figure}[t!]
  \centering
    \includegraphics[width=\columnwidth]{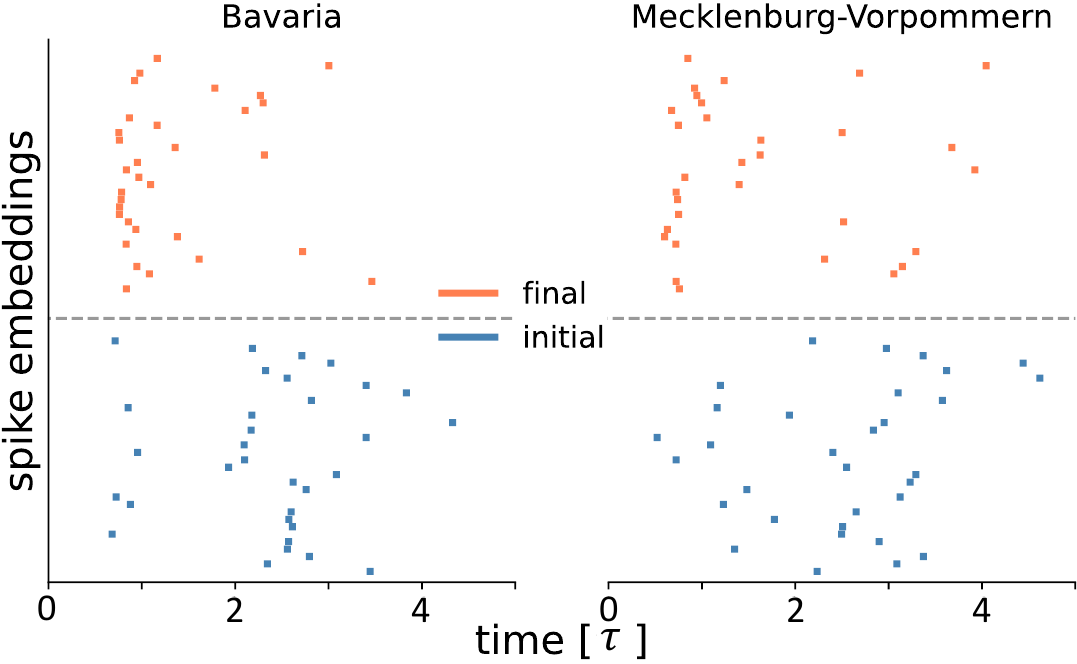}
    \Description{}
    \caption{Initial (bottom, blue) and final (top, orange) spike embeddings for two entities in a KG describing the geographical relationships of federal states in Germany. The ordinate shows the neurons of a single population, while the abscissa shows the respective spike times. Both the initial and final populations are active at the same time, with some neurons in the final embedding population even spiking before neurons in the initial embedding population. On average, we found that only 20-30\% of the initial embedding spikes are required to calculate the final embeddings. \copyright 2022 IEEE. Reprinted, with permission, from \cite{chian2021learning}.}\label{fig2}
\end{figure}

A normal GNN architecture with frozen weights achieves competitive results on benchmark tasks, e.g., a MRR of $0.80$ on UMLS and a MRR of $0.26$ on FB15k-237.
For the fully spiking system, due to long simulation times, we were only able to demonstrate that the model performs as well as other graph embedding models on smaller custom benchmarks.

Different from normal GNNs, our spike-based version calculates all iterations of the graph embeddings simultaneously.
Usually, the calculation of embeddings is locked layer-wise, i.e., the embedding of each node has to be updated before progressing to deeper layers in the GNN.
In our model, this is not the case: all layers operate in parallel, and final embeddings can be calculated even before some of the neurons in the initial layer have spiked, resulting in a novel and shallow way of encoding graph elements via spikes (see Fig.~\ref{fig2}).

\section{Conclusion}

We present a framework for symbolic computing with spiking neurons based on KGs and graph embedding algorithms.
Compared to previous approaches based on semantic pointers \cite{crawford2016biologically}, our method allows the learning of semantically meaningful, low-dimensional and purely spike-based representations of graph elements.
Due to the differentiability of our approach, such semantic representations can be trained end-to-end in unison with other differentiable models or architectures, such as multi-layer or recurrent SNNs, for specific use-cases, as demonstrated with our spiking GNN.
In addition, by changing the scoring function (Eq.~\ref{eq:spikedecoder}), alternative spike-based embedding schemes can be devised in future work that leverage gradient-based optimization in SNNs -- similarly to how a variety of embedding schemes exist in the graph embedding literature \cite{ruffinelli2019you}. 

The combination of symbolic data and spike-based computing is of particular interest for emerging neuromorphic technologies \cite{frenkel2021bottom}, as it bears the potential of opening up new data formats and applications.
Thus, we are convinced that our work constitutes an important step towards not only enabling, but also exploring neuro-symbolic reasoning with spiking systems.

\begin{acks}
We thank Serghei Mogoreanu, Alexander Hadjiivanov and Gabriele Meoni for helpful discussions.
We further thank our colleagues at the Semantics and Reasoning Research Group, the Siemens AI Lab and ESA's Advanced Concepts Team for their support.
This work was partially funded by the Federal Ministry for Economic Affairs and Energy of Germany (IIP-Ecosphere Project) and by the German Federal Ministry for Education and Research (“MLWin”, grant 01IS18050).
DD acknowledges support through the European Space Agency fellowship programme.
\end{acks}

\bibliographystyle{ACM-Reference-Format}
\bibliography{sample-base.bib}


\begin{thebibliography}{34}


\ifx \showCODEN    \undefined \def \showCODEN     #1{\unskip}     \fi
\ifx \showDOI      \undefined \def \showDOI       #1{#1}\fi
\ifx \showISBNx    \undefined \def \showISBNx     #1{\unskip}     \fi
\ifx \showISBNxiii \undefined \def \showISBNxiii  #1{\unskip}     \fi
\ifx \showISSN     \undefined \def \showISSN      #1{\unskip}     \fi
\ifx \showLCCN     \undefined \def \showLCCN      #1{\unskip}     \fi
\ifx \shownote     \undefined \def \shownote      #1{#1}          \fi
\ifx \showarticletitle \undefined \def \showarticletitle #1{#1}   \fi
\ifx \showURL      \undefined \def \showURL       {\relax}        \fi
\providecommand\bibfield[2]{#2}
\providecommand\bibinfo[2]{#2}
\providecommand\natexlab[1]{#1}
\providecommand\showeprint[2][]{arXiv:#2}

\bibitem[Ali and Kwisthout(2019)]%
        {ali2019spiking}
\bibfield{author}{\bibinfo{person}{Abdullahi Ali} {and} \bibinfo{person}{Johan
  Kwisthout}.} \bibinfo{year}{2019}\natexlab{}.
\newblock \showarticletitle{A spiking neural algorithm for the Network Flow
  problem}.
\newblock \bibinfo{journal}{\emph{arXiv:1911.13097}} (\bibinfo{year}{2019}).
\newblock


\bibitem[Auer et~al\mbox{.}(2007)]%
        {auer2007dbpedia}
\bibfield{author}{\bibinfo{person}{S{\"o}ren Auer}, \bibinfo{person}{Christian
  Bizer}, \bibinfo{person}{Georgi Kobilarov}, \bibinfo{person}{Jens Lehmann},
  \bibinfo{person}{Richard Cyganiak}, {and} \bibinfo{person}{Zachary Ives}.}
  \bibinfo{year}{2007}\natexlab{}.
\newblock \showarticletitle{Dbpedia: A nucleus for a web of open data}.
\newblock In \bibinfo{booktitle}{\emph{The semantic web}}.
  \bibinfo{publisher}{Springer}, \bibinfo{pages}{722--735}.
\newblock


\bibitem[Bollacker et~al\mbox{.}(2008)]%
        {bollacker2008freebase}
\bibfield{author}{\bibinfo{person}{Kurt Bollacker}, \bibinfo{person}{Colin
  Evans}, \bibinfo{person}{Praveen Paritosh}, \bibinfo{person}{Tim Sturge},
  {and} \bibinfo{person}{Jamie Taylor}.} \bibinfo{year}{2008}\natexlab{}.
\newblock \showarticletitle{Freebase: a collaboratively created graph database
  for structuring human knowledge}. In \bibinfo{booktitle}{\emph{Proceedings of
  the 2008 ACM SIGMOD international conference on Management of data}}.
  \bibinfo{pages}{1247--1250}.
\newblock


\bibitem[Bordes et~al\mbox{.}(2013)]%
        {bordes2013translating}
\bibfield{author}{\bibinfo{person}{Antoine Bordes}, \bibinfo{person}{Nicolas
  Usunier}, \bibinfo{person}{Alberto Garcia-Duran}, \bibinfo{person}{Jason
  Weston}, {and} \bibinfo{person}{Oksana Yakhnenko}.}
  \bibinfo{year}{2013}\natexlab{}.
\newblock \showarticletitle{Translating embeddings for modeling
  multi-relational data}. In \bibinfo{booktitle}{\emph{Advances in neural
  information processing systems}}. \bibinfo{pages}{2787--2795}.
\newblock


\bibitem[Brickley et~al\mbox{.}(1999)]%
        {brickley1999resource}
\bibfield{author}{\bibinfo{person}{Dan Brickley}, \bibinfo{person}{Ramanathan~V
  Guha}, {and} \bibinfo{person}{Andrew Layman}.}
  \bibinfo{year}{1999}\natexlab{}.
\newblock \showarticletitle{Resource description framework (RDF) schema
  specification}.
\newblock \bibinfo{journal}{\emph{Technical report, W3C.}}
  (\bibinfo{year}{1999}).
\newblock


\bibitem[Chian et~al\mbox{.}(2021)]%
        {chian2021learning}
\bibfield{author}{\bibinfo{person}{Victor~Caceres Chian},
  \bibinfo{person}{Marcel Hildebrandt}, \bibinfo{person}{Thomas Runkler}, {and}
  \bibinfo{person}{Dominik Dold}.} \bibinfo{year}{2021}\natexlab{}.
\newblock \showarticletitle{Learning through structure: towards deep
  neuromorphic knowledge graph embeddings}. In \bibinfo{booktitle}{\emph{2021
  International Conference on Neuromorphic Computing (ICNC)}}. IEEE,
  \bibinfo{pages}{61--70}.
\newblock


\bibitem[Comsa et~al\mbox{.}(2020)]%
        {comsa2019temporal}
\bibfield{author}{\bibinfo{person}{Iulia~M Comsa}, \bibinfo{person}{Thomas
  Fischbacher}, \bibinfo{person}{Krzysztof Potempa}, \bibinfo{person}{Andrea
  Gesmundo}, \bibinfo{person}{Luca Versari}, {and} \bibinfo{person}{Jyrki
  Alakuijala}.} \bibinfo{year}{2020}\natexlab{}.
\newblock \showarticletitle{Temporal coding in spiking neural networks with
  alpha synaptic function}. In \bibinfo{booktitle}{\emph{IEEE International
  Conference on Acoustics, Speech and Signal Processing (ICASSP)}}. IEEE,
  \bibinfo{pages}{8529--8533}.
\newblock


\bibitem[Crawford et~al\mbox{.}(2016)]%
        {crawford2016biologically}
\bibfield{author}{\bibinfo{person}{Eric Crawford}, \bibinfo{person}{Matthew
  Gingerich}, {and} \bibinfo{person}{Chris Eliasmith}.}
  \bibinfo{year}{2016}\natexlab{}.
\newblock \showarticletitle{Biologically plausible, human-scale knowledge
  representation}.
\newblock \bibinfo{journal}{\emph{Cognitive science}} \bibinfo{volume}{40},
  \bibinfo{number}{4} (\bibinfo{year}{2016}), \bibinfo{pages}{782--821}.
\newblock


\bibitem[Dold(2022)]%
        {dold2022spikeembed}
\bibfield{author}{\bibinfo{person}{Dominik Dold}.}
  \bibinfo{year}{2022}\natexlab{}.
\newblock \showarticletitle{Relational representation learning with spike
  trains}.
\newblock \bibinfo{journal}{\emph{International Joint Conference on Neural
  Networks (IJCNN)}} (\bibinfo{year}{2022}).
\newblock


\bibitem[Dold and Soler-Garrido(2021)]%
        {dold2021spikeembed}
\bibfield{author}{\bibinfo{person}{Dominik Dold} {and} \bibinfo{person}{Josep
  Soler-Garrido}.} \bibinfo{year}{2021}\natexlab{}.
\newblock \showarticletitle{SpikE: spike-based embeddings for multi-relational
  graph data}.
\newblock \bibinfo{journal}{\emph{International Joint Conference on Neural
  Networks (IJCNN)}} (\bibinfo{year}{2021}).
\newblock


\bibitem[Frenkel et~al\mbox{.}(2021)]%
        {frenkel2021bottom}
\bibfield{author}{\bibinfo{person}{Charlotte Frenkel}, \bibinfo{person}{David
  Bol}, {and} \bibinfo{person}{Giacomo Indiveri}.}
  \bibinfo{year}{2021}\natexlab{}.
\newblock \showarticletitle{Bottom-Up and Top-Down Neural Processing Systems
  Design: Neuromorphic Intelligence as the Convergence of Natural and
  Artificial Intelligence}.
\newblock \bibinfo{journal}{\emph{arXiv preprint arXiv:2106.01288}}
  (\bibinfo{year}{2021}).
\newblock


\bibitem[G{\"o}ltz et~al\mbox{.}(2021)]%
        {goltz2021fast}
\bibfield{author}{\bibinfo{person}{Julian G{\"o}ltz}, \bibinfo{person}{L
  Kriener}, \bibinfo{person}{A Baumbach}, \bibinfo{person}{S Billaudelle},
  \bibinfo{person}{O Breitwieser}, \bibinfo{person}{B Cramer},
  \bibinfo{person}{D Dold}, \bibinfo{person}{AF Kungl}, \bibinfo{person}{W
  Senn}, \bibinfo{person}{J Schemmel}, {et~al\mbox{.}}}
  \bibinfo{year}{2021}\natexlab{}.
\newblock \showarticletitle{Fast and energy-efficient neuromorphic deep
  learning with first-spike times}.
\newblock \bibinfo{journal}{\emph{Nature Machine Intelligence}}
  \bibinfo{volume}{3}, \bibinfo{number}{9} (\bibinfo{year}{2021}),
  \bibinfo{pages}{823--835}.
\newblock


\bibitem[Hamilton et~al\mbox{.}(2017a)]%
        {hamilton2017community}
\bibfield{author}{\bibinfo{person}{Kathleen~E Hamilton}, \bibinfo{person}{Neena
  Imam}, {and} \bibinfo{person}{Travis~S Humble}.}
  \bibinfo{year}{2017}\natexlab{a}.
\newblock \showarticletitle{Community detection with spiking neural networks
  for neuromorphic hardware}. In \bibinfo{booktitle}{\emph{Proceedings of the
  Neuromorphic Computing Symposium}}. \bibinfo{pages}{1--8}.
\newblock


\bibitem[Hamilton et~al\mbox{.}(2019)]%
        {hamilton2019spike}
\bibfield{author}{\bibinfo{person}{Kathleen~E Hamilton},
  \bibinfo{person}{Tiffany~M Mintz}, {and} \bibinfo{person}{Catherine~D
  Schuman}.} \bibinfo{year}{2019}\natexlab{}.
\newblock \showarticletitle{Spike-based primitives for graph algorithms}.
\newblock \bibinfo{journal}{\emph{arXiv:1903.10574}} (\bibinfo{year}{2019}).
\newblock


\bibitem[Hamilton and Schuman(2018)]%
        {hamilton2018towards}
\bibfield{author}{\bibinfo{person}{Kathleen~E Hamilton} {and}
  \bibinfo{person}{Catherine~D Schuman}.} \bibinfo{year}{2018}\natexlab{}.
\newblock \showarticletitle{Towards adaptive spiking label propagation}. In
  \bibinfo{booktitle}{\emph{Proceedings of the International Conference on
  Neuromorphic Systems}}. \bibinfo{pages}{1--8}.
\newblock


\bibitem[Hamilton et~al\mbox{.}(2018)]%
        {hamilton2018neural}
\bibfield{author}{\bibinfo{person}{Kathleen~E Hamilton},
  \bibinfo{person}{Catherine~D Schuman}, \bibinfo{person}{Steven~R Young},
  \bibinfo{person}{Neena Imam}, {and} \bibinfo{person}{Travis~S Humble}.}
  \bibinfo{year}{2018}\natexlab{}.
\newblock \showarticletitle{Neural networks and graph algorithms with
  next-generation processors}. In \bibinfo{booktitle}{\emph{2018 IEEE
  International Parallel and Distributed Processing Symposium Workshops
  (IPDPSW)}}. IEEE, \bibinfo{pages}{1194--1203}.
\newblock


\bibitem[Hamilton et~al\mbox{.}(2017b)]%
        {hamilton2017inductive}
\bibfield{author}{\bibinfo{person}{Will Hamilton}, \bibinfo{person}{Zhitao
  Ying}, {and} \bibinfo{person}{Jure Leskovec}.}
  \bibinfo{year}{2017}\natexlab{b}.
\newblock \showarticletitle{Inductive representation learning on large graphs}.
\newblock \bibinfo{journal}{\emph{Advances in neural information processing
  systems}}  \bibinfo{volume}{30} (\bibinfo{year}{2017}).
\newblock


\bibitem[Hamilton et~al\mbox{.}(2017c)]%
        {hamilton2017representation}
\bibfield{author}{\bibinfo{person}{William~L Hamilton}, \bibinfo{person}{Rex
  Ying}, {and} \bibinfo{person}{Jure Leskovec}.}
  \bibinfo{year}{2017}\natexlab{c}.
\newblock \showarticletitle{Representation learning on graphs: Methods and
  applications}.
\newblock \bibinfo{journal}{\emph{IEEE Data Engineering Bulletin}}
  (\bibinfo{year}{2017}).
\newblock


\bibitem[Kay et~al\mbox{.}(2020)]%
        {kay2020neuromorphic}
\bibfield{author}{\bibinfo{person}{Bill Kay}, \bibinfo{person}{Prasanna Date},
  {and} \bibinfo{person}{Catherine Schuman}.} \bibinfo{year}{2020}\natexlab{}.
\newblock \showarticletitle{Neuromorphic graph algorithms: Extracting longest
  shortest paths and minimum spanning trees}. In
  \bibinfo{booktitle}{\emph{Proceedings of the Neuro-inspired Computational
  Elements Workshop}}. \bibinfo{pages}{1--6}.
\newblock


\bibitem[Kheradpisheh and Masquelier(2020)]%
        {kheradpisheh2019s4nn}
\bibfield{author}{\bibinfo{person}{Saeed~Reza Kheradpisheh} {and}
  \bibinfo{person}{Timoth{\'e}e Masquelier}.} \bibinfo{year}{2020}\natexlab{}.
\newblock \showarticletitle{S4NN: temporal backpropagation for spiking neural
  networks with one spike per neuron}.
\newblock \bibinfo{journal}{\emph{International Journal of Neural Systems}}
  \bibinfo{volume}{30}, \bibinfo{number}{6} (\bibinfo{year}{2020}),
  \bibinfo{pages}{2050027}.
\newblock


\bibitem[Kipf and Welling(2016)]%
        {kipf2016semi}
\bibfield{author}{\bibinfo{person}{Thomas~N Kipf} {and} \bibinfo{person}{Max
  Welling}.} \bibinfo{year}{2016}\natexlab{}.
\newblock \showarticletitle{Semi-supervised classification with graph
  convolutional networks}.
\newblock \bibinfo{journal}{\emph{arXiv preprint arXiv:1609.02907}}
  (\bibinfo{year}{2016}).
\newblock


\bibitem[Lehmann(2009)]%
        {lehmann2009dl}
\bibfield{author}{\bibinfo{person}{Jens Lehmann}.}
  \bibinfo{year}{2009}\natexlab{}.
\newblock \showarticletitle{DL-Learner: learning concepts in description
  logics}.
\newblock \bibinfo{journal}{\emph{The Journal of Machine Learning Research}}
  \bibinfo{volume}{10} (\bibinfo{year}{2009}), \bibinfo{pages}{2639--2642}.
\newblock


\bibitem[McCray(2003)]%
        {mccray2003upper}
\bibfield{author}{\bibinfo{person}{Alexa~T McCray}.}
  \bibinfo{year}{2003}\natexlab{}.
\newblock \showarticletitle{An upper-level ontology for the biomedical domain}.
\newblock \bibinfo{journal}{\emph{Comparative and Functional genomics}}
  \bibinfo{volume}{4}, \bibinfo{number}{1} (\bibinfo{year}{2003}),
  \bibinfo{pages}{80--84}.
\newblock


\bibitem[Mostafa(2017)]%
        {mostafa2017supervised}
\bibfield{author}{\bibinfo{person}{Hesham Mostafa}.}
  \bibinfo{year}{2017}\natexlab{}.
\newblock \showarticletitle{Supervised learning based on temporal coding in
  spiking neural networks}.
\newblock \bibinfo{journal}{\emph{IEEE transactions on neural networks and
  learning systems}} \bibinfo{volume}{29}, \bibinfo{number}{7}
  (\bibinfo{year}{2017}), \bibinfo{pages}{3227--3235}.
\newblock


\bibitem[Neftci et~al\mbox{.}(2019)]%
        {neftci2019surrogate}
\bibfield{author}{\bibinfo{person}{Emre~O Neftci}, \bibinfo{person}{Hesham
  Mostafa}, {and} \bibinfo{person}{Friedemann Zenke}.}
  \bibinfo{year}{2019}\natexlab{}.
\newblock \showarticletitle{Surrogate gradient learning in spiking neural
  networks: Bringing the power of gradient-based optimization to spiking neural
  networks}.
\newblock \bibinfo{journal}{\emph{IEEE Signal Processing Magazine}}
  \bibinfo{volume}{36}, \bibinfo{number}{6} (\bibinfo{year}{2019}),
  \bibinfo{pages}{51--63}.
\newblock


\bibitem[Nickel et~al\mbox{.}(2015)]%
        {nickel2015review}
\bibfield{author}{\bibinfo{person}{Maximilian Nickel}, \bibinfo{person}{Kevin
  Murphy}, \bibinfo{person}{Volker Tresp}, {and} \bibinfo{person}{Evgeniy
  Gabrilovich}.} \bibinfo{year}{2015}\natexlab{}.
\newblock \showarticletitle{A review of relational machine learning for
  knowledge graphs}.
\newblock \bibinfo{journal}{\emph{Proc. IEEE}} \bibinfo{volume}{104},
  \bibinfo{number}{1} (\bibinfo{year}{2015}), \bibinfo{pages}{11--33}.
\newblock


\bibitem[Ruffinelli et~al\mbox{.}(2019)]%
        {ruffinelli2019you}
\bibfield{author}{\bibinfo{person}{Daniel Ruffinelli}, \bibinfo{person}{Samuel
  Broscheit}, {and} \bibinfo{person}{Rainer Gemulla}.}
  \bibinfo{year}{2019}\natexlab{}.
\newblock \showarticletitle{You CAN teach an old dog new tricks! on training
  knowledge graph embeddings}. In \bibinfo{booktitle}{\emph{International
  Conference on Learning Representations}}.
\newblock


\bibitem[Schlichtkrull et~al\mbox{.}(2018)]%
        {schlichtkrull2018modeling}
\bibfield{author}{\bibinfo{person}{Michael Schlichtkrull},
  \bibinfo{person}{Thomas~N Kipf}, \bibinfo{person}{Peter Bloem},
  \bibinfo{person}{Rianne van~den Berg}, \bibinfo{person}{Ivan Titov}, {and}
  \bibinfo{person}{Max Welling}.} \bibinfo{year}{2018}\natexlab{}.
\newblock \showarticletitle{Modeling relational data with graph convolutional
  networks}. In \bibinfo{booktitle}{\emph{European semantic web conference}}.
  Springer, \bibinfo{pages}{593--607}.
\newblock


\bibitem[Schuman et~al\mbox{.}(2019)]%
        {schuman2019shortest}
\bibfield{author}{\bibinfo{person}{Catherine~D Schuman},
  \bibinfo{person}{Kathleen Hamilton}, \bibinfo{person}{Tiffany Mintz},
  \bibinfo{person}{Md~Musabbir Adnan}, \bibinfo{person}{Bon~Woong Ku},
  \bibinfo{person}{Sung-Kyu Lim}, {and} \bibinfo{person}{Garrett~S Rose}.}
  \bibinfo{year}{2019}\natexlab{}.
\newblock \showarticletitle{Shortest path and neighborhood subgraph extraction
  on a spiking memristive neuromorphic implementation}. In
  \bibinfo{booktitle}{\emph{Proceedings of the 7th Annual Neuro-inspired
  Computational Elements Workshop}}. \bibinfo{pages}{1--6}.
\newblock


\bibitem[Singhal(2012)]%
        {singhal2012introducing}
\bibfield{author}{\bibinfo{person}{Amit Singhal}.}
  \bibinfo{year}{2012}\natexlab{}.
\newblock \showarticletitle{Introducing the knowledge graph: things, not
  strings, May 2012}.
\newblock \bibinfo{journal}{\emph{URL http://googleblog. blogspot.
  ie/2012/05/introducing-knowledgegraph-things-not. html}}
  (\bibinfo{year}{2012}).
\newblock


\bibitem[Wunderlich and Pehle(2021)]%
        {wunderlich2021event}
\bibfield{author}{\bibinfo{person}{Timo~C Wunderlich} {and}
  \bibinfo{person}{Christian Pehle}.} \bibinfo{year}{2021}\natexlab{}.
\newblock \showarticletitle{Event-based backpropagation can compute exact
  gradients for spiking neural networks}.
\newblock \bibinfo{journal}{\emph{Scientific Reports}} \bibinfo{volume}{11},
  \bibinfo{number}{1} (\bibinfo{year}{2021}), \bibinfo{pages}{1--17}.
\newblock


\bibitem[Yin et~al\mbox{.}(2021)]%
        {yin2021accurate}
\bibfield{author}{\bibinfo{person}{Bojian Yin}, \bibinfo{person}{Federico
  Corradi}, {and} \bibinfo{person}{Sander~M Boht{\'e}}.}
  \bibinfo{year}{2021}\natexlab{}.
\newblock \showarticletitle{Accurate and efficient time-domain classification
  with adaptive spiking recurrent neural networks}.
\newblock \bibinfo{journal}{\emph{Nature Machine Intelligence}}
  \bibinfo{volume}{3}, \bibinfo{number}{10} (\bibinfo{year}{2021}),
  \bibinfo{pages}{905--913}.
\newblock


\bibitem[Zenke et~al\mbox{.}(2021)]%
        {zenke2021visualizing}
\bibfield{author}{\bibinfo{person}{Friedemann Zenke}, \bibinfo{person}{Sander~M
  Boht{\'e}}, \bibinfo{person}{Claudia Clopath}, \bibinfo{person}{Iulia~M
  Com{\c{s}}a}, \bibinfo{person}{Julian G{\"o}ltz}, \bibinfo{person}{Wolfgang
  Maass}, \bibinfo{person}{Timoth{\'e}e Masquelier}, \bibinfo{person}{Richard
  Naud}, \bibinfo{person}{Emre~O Neftci}, \bibinfo{person}{Mihai~A Petrovici},
  {et~al\mbox{.}}} \bibinfo{year}{2021}\natexlab{}.
\newblock \showarticletitle{Visualizing a joint future of neuroscience and
  neuromorphic engineering}.
\newblock \bibinfo{journal}{\emph{Neuron}} \bibinfo{volume}{109},
  \bibinfo{number}{4} (\bibinfo{year}{2021}), \bibinfo{pages}{571--575}.
\newblock


\bibitem[Zenke and Ganguli(2018)]%
        {zenke2018superspike}
\bibfield{author}{\bibinfo{person}{Friedemann Zenke} {and}
  \bibinfo{person}{Surya Ganguli}.} \bibinfo{year}{2018}\natexlab{}.
\newblock \showarticletitle{SuperSpike: Supervised Learning in Multilayer
  Spiking Neural Networks}.
\newblock \bibinfo{journal}{\emph{Neural computation}} \bibinfo{volume}{30},
  \bibinfo{number}{6} (\bibinfo{year}{2018}), \bibinfo{pages}{1514--1541}.
\newblock


\end{thebibliography}
\end{document}